\documentclass[journal]{IEEEtran}

\usepackage{cite}

\ifCLASSINFOpdf
  \usepackage[pdftex]{graphicx}
  \graphicspath{{../pdf/}{../jpeg/}}
  \DeclareGraphicsExtensions{.pdf,.jpeg,.png}
\else
  \usepackage[dvips]{graphicx}
  \graphicspath{{../eps/}}
  \DeclareGraphicsExtensions{.eps}
\fi

\usepackage{amsmath,amssymb}
\usepackage{relsize}
\def\bv{{\mathbf v}}

\usepackage{algorithmic}

\usepackage{array}
\usepackage{multirow}
\newcolumntype{P}[1]{>{\centering\arraybackslash}p{#1}}
\newcolumntype{M}[1]{>{\centering\arraybackslash}m{#1}}

\ifCLASSOPTIONcompsoc
  \usepackage[caption=false,font=normalsize,labelfont=sf,textfont=sf]{subfig}
\else
  \usepackage[caption=false,font=footnotesize]{subfig}
\fi

\usepackage{fixltx2e}
\ifCLASSOPTIONcaptionsoff
  \usepackage[nomarkers]{endfloat}
 \let\MYoriglatexcaption\caption
 \renewcommand{\caption}[2][\relax]{\MYoriglatexcaption[#2]{#2}}
\fi

\usepackage{url}
\usepackage{amsmath,amsthm}
\usepackage{amsfonts}
\usepackage{amssymb}
\usepackage{multirow}
\usepackage[pdftex]{graphicx}
\usepackage{cite}
\usepackage{hyperref}
\hypersetup{colorlinks=true,urlcolor=red}
\usepackage{tabularx}
\usepackage{array}

\begin{document}

\title{CycleMorph: Cycle Consistent Unsupervised Deformable Image Registration}

\author{Boah~Kim, Dong~Hwan~Kim, Seong~Ho~Park, Jieun~Kim, June-Goo~Lee, and~Jong~Chul~Ye,~\IEEEmembership{Fellow,~IEEE}
\thanks{B. Kim and J.C. Ye are with the Department of Bio and Brain Engineering, Korea Advanced Institute of Science and Technology (KAIST), Daejeon 34141, Republic of Korea (email: \{boahkim, jong.ye\}@kaist.ac.kr). 

D.H. Kim is with the Department of Radiology, Seoul St. Mary's Hospital, College of Medicine, The Catholic University of Korea, Seoul, Republic of Korea.

S.H. Park is with the Department of Radiology and Research Institute of Radiology, University of Ulsan College of Medicine, Asan Medical Center, Seoul, Republic of Korea.

J. Kim is with the Smart Car R\&D Division, AI-Bigdata R\&D Center, Korea Automotive Technology Institute (KATECH), Republic of Korea (email: kimje@katech.re.kr).

J.G. Lee is with the Department of Convergence Medicine, Asan Medical Institute of Convergence Science and Technology, Asan Medical Center, University of Ulsan College of Medicine, Seoul, Republic of Korea.

Part of this work was presented at the Medical Image Computing and Computer Assisted Intervention (MICCAI) 2019 conference \cite{kim2019unsupervised}. 

This work was supported in part by the Industrial Strategic technology development program (10072064, Development of Novel Artificial Intelligence Technologies To Assist Imaging Diagnosis of Pulmonary, Hepatic, and Cardiac Diseases and Their Integration into Commercial Clinical PACS Platforms) funded by the Ministry of Trade Industry and Energy (MI, Korea), and also supported in part by KAIST R\&D Program (KI Meta-Convergence Program) 2020 through Korea Advanced Institute of Science and Technology (KAIST).}
}%

%
%

%


\maketitle

\begin{abstract}
Image registration is a fundamental task in medical image analysis. 
Recently, deep learning based image registration methods have been extensively investigated due to their excellent performance despite the ultra-fast computational time. However, the existing deep learning methods still have limitation in the preservation of original topology during the deformation with registration vector fields. To address this issues, here we present a cycle-consistent deformable image registration. The cycle consistency enhances image registration performance by providing an implicit  regularization to preserve topology during the deformation. The proposed method is so flexible that can be applied for both 2D and 3D registration problems for various applications, and can be easily extended to multi-scale implementation to deal with the memory issues in large volume registration. 
Experimental results on various datasets from medical and non-medical applications demonstrate that the proposed method provides effective and accurate registration on diverse image pairs within a few seconds. Qualitative and quantitative evaluations on deformation fields also verify the effectiveness of the cycle consistency of the proposed method.

\end{abstract}


%
\IEEEpeerreviewmaketitle

\section{Introduction}

\begin{figure}[t!]
\centering
\includegraphics[width=8.8cm]{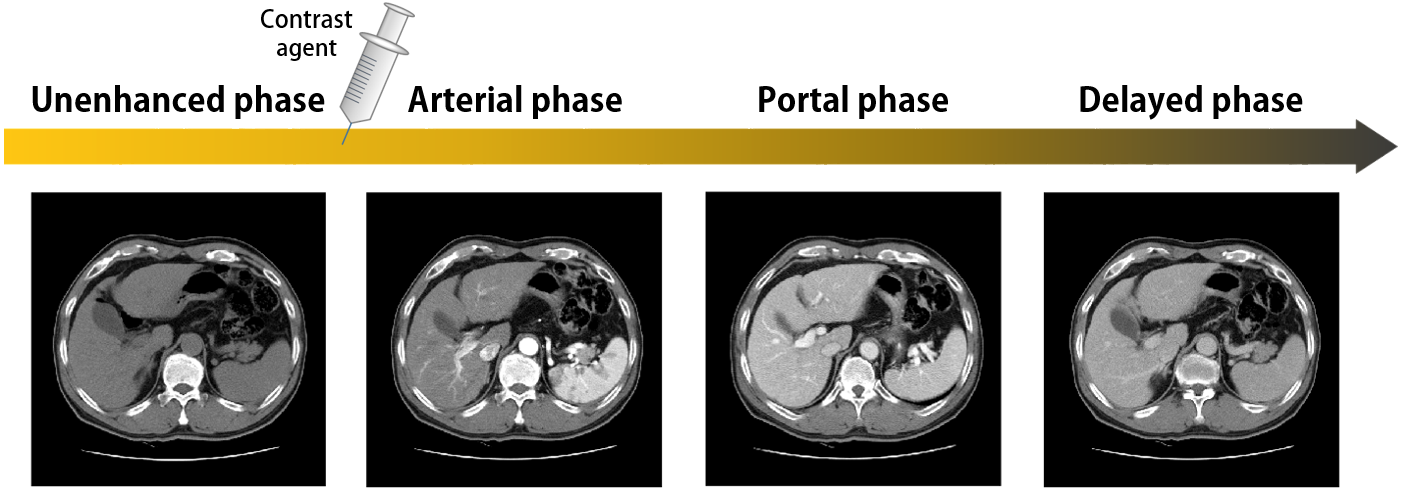}
\caption{Example of 2D slices taken from 3D liver CT volumes before and after injection of contrast agent. Images at different phases show various contrast and shape of liver and other organs.}
\label{fig:introExample}
\end{figure}

\IEEEPARstart{I}{mage} registration is one of the fundamental tasks in medical imaging, since the shape of anatomical structures in images vary due to the disease progress, patient motion, imaging modalities, etc. 
For example,  radiologists often diagnose the liver tumor with multiphase contrast enhanced CT (CECT) images \cite{kim2011assessment}, but the images at different temporal phases are usually different in their shape and image contrast as shown in Fig. \ref{fig:introExample}.  
 
 Although conventional image registration methods \cite{christensen2001consistent, leow2005inverse, ashburner2007fast, beg2005computing, avants2008symmetric, klein2009elastix} have been studied to address this using a variational framework that solves an optimization problem for each image pair to be aligned with similar appearance, these approaches usually suffer from  extensive computation and long registration time.

Recently, deep learning approaches have demonstrated performance improvement over the traditional methods for image registration. Given source and target images, deep neural networks are trained to generate deformation fields corresponding to the input image pair, so that it enables significant fast registration \cite{onofrey2013semi, zhang2008rapid, shang2006rigid, yang2017quicksilver, rohe2017svf, sokooti2017nonrigid}. Nowadays,  these methods have been evolved to unsupervised learning methods that do not require ground-truth deformation fields \cite{krebs2018learning, balakrishnan2018unsupervised, fan2018adversarial, lei20204d}. However, the existing image registration approaches do not explicitly enforce the criterion to guarantee topology preservation, which often result in inaccurate registration with the loss of structural information.

To overcome the potential degeneracy problem of registration field, here we present a novel deformable image registration method called CycleMorph, which uses cycle consistency to force the deformed image to return to the original image  \cite{zhu2017unpaired}.
In contrast to the existing approaches that enforces the inverse-consistency to the deformation vector fields generated from additional inverse networks \cite{zhang2018inverse}, one of the most important contributions of this work is the demonstration of the topological preservation by imposing the cycle consistency simply on the images.  

More specifically, we train two convolutional neural networks (CNN), $G_X$ and $G_Y$, that generate forward and reverse directional deformation vector fields, respectively. When a moving source image is deformed to the other fixed image by the deformation field from $G_X$, then the deformed image can be reversed to the original image using the deformation field from $G_Y$, by applying the cycle consistency to the reversed image and the original image. It turns out that this inverse path with the cyclic constraint is a direct way of providing high performance topology preservation with less folding problem during the deformation process.

Another important innovation of this work is the extension to multi-scale implementation to deal with the large volume image registration. Specifically, due to the GPU memory limitation, training with the whole 3D volume for image registration may not be possible. To deal with this, we propose a coarse 3D registration using the subsampled volume for large deformation, followed by local deformation estimation to improve the registration accuracy.

In order to verify the performance of the proposed method, we apply our algorithm to  various applications from different domains with varying memory requirement, including 2D face registration, 3D brain MR registration,  and multiphase 3D abdominal contrast enhanced CT (CECT) volume registration for liver cancer evaluation. Qualitative and quantitative evaluation of the experimental results demonstrate the robustness of the proposed method and confirm the efficacy of the cycle consistency for topology preservation.

The paper is organized as follows. Section~\ref{sec:related} review the related works. 
Section~\ref{sec:method} describes our theory and proposed method. Section~\ref{sec:experiment} presents experimental results and discussion on registration of face expression image, MRI, and CT dataset, and we conclude in Section~\ref{sec:conclusion}.

\section{Related Works}
\label{sec:related}
\subsection{Diffeomorphic Image Registration}

In classical variational image registration approaches, an energy function is typically composed of two terms: 
 \begin{align}
\mathcal{L}(X, Y, {\phi}) 
=&  \mathcal{L}_{sim}\left(\mathcal{T}(X, \phi), Y\right) +  \mathcal{L}_{reg}({\phi})
\label{eq:optimization_problem}
\end{align}
where $X$ and $Y$ denote the moving image and fixed image, respectively; $\phi$ represents the displacement vector field, and $\mathcal{T}$ is the transformation function which warps $X$ to $Y$ using the deformation vector field $\phi$. In \eqref{eq:optimization_problem}, the first term is a similarity function which evaluates the shape differences between deformed images and reference images, whereas the second term is a regularization function to make the deformation field smooth. 

In particular, diffeomorphic image registration methods imposes the constraint on the vector field $\phi$ such that the resulting deformable mapping becomes a diffeomorphism. A diffeomorphic deformation ensures certain desirable properties between two image volumes like continuous, differentiable, and preserving topology \cite{beg2005computing, avants2008symmetric, vercauteren2009diffeomorphic}. The popular examples of these algorithmic extensions to large deformation are Large Deformation Diffeomorphic Metric Matching (LDDMM) \cite{beg2005computing, zhang2017frequency, cao2005large, ceritoglu2009multi} and Symmetric image Normalization method (SyN) \cite{avants2008symmetric}. 

Unfortunately, these algorithms are usually computationally expensive, which prohibits its routine use in clinical workflow.

\subsection{Deep-learning-based Image Registration}

On the other hand, the learning-based registration algorithms are inductive  in the sense that once a neural network is trained, it  can instantaneously  predict deformation vector fields for a new data. Therefore, it is ideally suitable for clinical environment.
Depending on how the networks are trained, the methods are categorized into two types: supervised learning methods and unsupervised learning methods. In the following, we provide more details.

\subsubsection{Supervised Learning Methods}
In supervised learning approaches, the ground-truth of deformation vector fields are required, which are typically generated by the classical registration methods \cite{cao2017deformable, yang2017quicksilver, cao2018deep, rohe2017svf, sokooti2017nonrigid}. Yang et al. \cite{yang2017quicksilver} proposed an encoder-decoder network for patch-wise prediction of the deformation field, and a correction network to improve deformation prediction. Cao et al. \cite{cao2018deep} developed a non-rigid inter-modality image registration network that estimates registration fields of two-modal images. However, since the registration performance of these approaches depends on quality of the ground-truths, these works often require high quality ground-truth deformation fields and complicated pre-processing, both of which are often difficult to obtain in practice.

\begin{figure*}[t!]
\centering
\includegraphics[width=\linewidth]{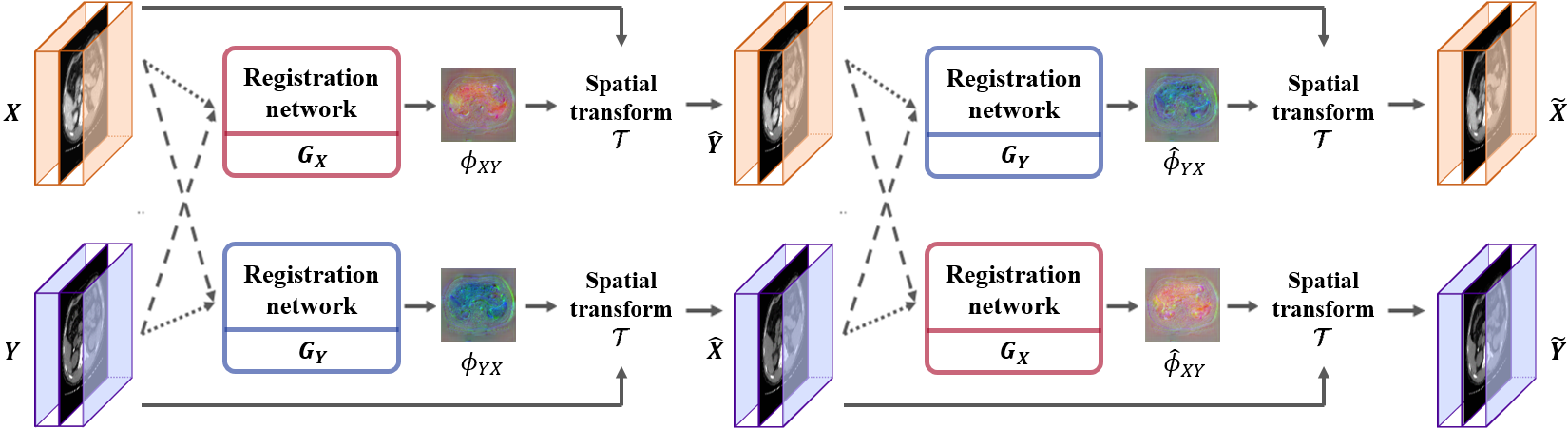}
\caption{The overall framework of the proposed cycle consistent deep learning model, CycleMorph, for deformable image registration. Two registration networks ($G_X, G_Y$) are used to take inputs by switching their order. Each networks takes two volumes ($X, Y$) and computes displacement vector field with three channels. Short and long dashed lines denote the moving images and fixed images, respectively. The spatial transform function deforms the moving image according to the vector field to match a shape of the fixed image. These transformed images ($\hat{X}, \hat{Y}$) are taken to the networks followed by transform function to ensure that the deformed images can be returned to original state.}
\label{fig:pipeline}
\end{figure*}

\subsubsection{Unsupervised Learning Methods}
To overcome the limitation of supervised learning approaches, unsupervised learning methods have been recently developed, which learn the image registration  by minimizing the loss between the deformed image and fixed target image.
Kreb et al. \cite{krebs2018learning} proposed an unsupervised learning model of a low-dimensional stochastic parametrization of the deformation by minimizing the KL divergence between two image distributions. Balakrishnan et al. \cite{balakrishnan2018unsupervised,balakrishnan2019voxelmorph} presented a pairwise 3D medical image registration algorithm using a CNN with a spatial transform layer (STL), which parameters are learned by the normalized cross-correlation function. 
For large volume image registration, Vos et al. \cite{de2019deep} proposed affine and non-rigid image registration framework, and Lei et al. \cite{lei20204d} presented a multiscale unsupervised learning method called MS-DIRNet through global and local registration networks. 

However, these methods do not usually impose the constraint for the consistency, so that they can cause a folding problem from the degeneracy of the mapping. Although Dalca et al. \cite{dalca2018unsupervised} introduced a diffeomorphic integration layers to address this issue, the constraint should
be also applied at the inference phase, which incurs additional complexity.


\subsection{Consistent Image Registration}
Although the classical diffeomorphic deformable registration algorithms have been proposed to ensure the one-to-one correspondence, deformations are generally represented discretely with a finite number of parameters, so there may be some small violations. Thus, the estimated deformation $F: X \mapsto Y$  is not equal to the inverse of the estimated deformation from $R: Y \mapsto X$. In consistent image registration approaches \cite{christensen2001consistent, ashburner2007fast, leow2005inverse},  this problem is alleviated by imposing additional inverse consistency
\begin{eqnarray}\label{eq:inv}
R\simeq F^{-1} \ .
\end{eqnarray}
In particular, the forward and inverse mappings $F$ and $R$ are only defined through the corresponding deformation fields $\phi_{XY}$ and $\phi_{YX}$, so the corresponding inverse-consistency is usually enforced as a regularization term to the deformation vector fields.

Recently, Zhang \cite{zhang2018inverse} proposed an inverse-consistency enforced deep learning model that simultaneously trains both forward and inverse neural networks. The forward network estimates the deformation field that can map a source to the target, whereas the inverse network generates the inverse flow under the inverse consistency condition of the deformation fields. On the other hand, Mahapatra et al. \cite{mahapatra2018deformable} proposed GAN-based image registration method by exploiting cycle consistency \cite{zhu2017unpaired} on the deformed images. However, this method should have pairs of perfectly landmarks-aligned images for network training.

\section{Theory}
\label{sec:method}
The overall learning framework of the proposed CycleMorph is illustrated in Fig. \ref{fig:pipeline}. Specifically, for the moving source and fixed target images, $X$ and $Y$, in different shapes or contrast, we define two registration networks as $G_{X}:(X, Y) \rightarrow \phi_{XY}$ and $G_{Y}:(Y, X) \rightarrow \phi_{YX}$, where $\phi_{XY}$ (resp. $\phi_{YX}$) denotes the  deformation field from $X$ to $Y$ (resp. $Y$ to $X$). We use a spatial transformation layer $\mathcal{T}$ in the networks to warp the moving image by the estimated deformation field, so that the registration networks are trained to minimize the dissimilarity between the deformed image and fixed image. Accordingly, when a pair of images are given to the registration networks, the moving image is deformed to align with the fixed image.

In particular, to guarantee the topology preservation between the deformed and fixed images, we employ the cycle consistency constraint between the original moving image and its re-deformed image. That is, the two deformed images are given as an input to the networks again by switching their order to impose cycle consistency on a pixel level of images. This constraint allows the networks to provide diffeomorphic deformation by ensuring the shape of deformed images successively return to the original shape.

\subsection{Loss Function}
We train the proposed cycle consistent learning model by solving the following optimization problem:
\begin{eqnarray}
\min_{G_X,G_Y} \mathcal{L}(X, Y, G_X, G_Y),
\end{eqnarray}
where 
\begin{align}
\mathcal{L}(X, Y, G_X, G_Y) 
=&  \mathcal{L}_{regist}(X, Y, {G_X}) \nonumber \\
&+  \mathcal{L}_{regist}(Y, X, {G_Y}) \nonumber \\
&+ \alpha \mathcal{L}_{cycle}(X, Y, {G_X}, {G_Y}) \nonumber \\
&+ \beta \mathcal{L}_{identity}(X, Y, {G_X}, {G_Y}), 
\label{eq:loss}
\end{align}
where $\mathcal{L}_{regist}$, $\mathcal{L}_{cycle}$, and $\mathcal{L}_{identity}$ are registration loss, cycle loss, and identity loss, respectively, and $\alpha$ and $\beta$ are hyper-parameters. As shown in Fig.~\ref{fig:loss}, our method is trained in an unsupervised manner without ground-truth deformation fields. More detailed description of each loss functions is as following.

\subsubsection{Registration Loss}
The registration loss function is based on the energy function of traditional variational image registration~\eqref{eq:optimization_problem} that has similarity and smoothness penalized terms. We employ the local cross-correlation for the similarity function to be less sensitive to the contrast variations \cite{avants2008symmetric}, and the $l_2$-loss for the regularization function. Accordingly, our registration loss function can be written as:
\begin{align}
\mathcal{L}_{regist} (X, & Y, G_X) \nonumber \\
& =-(\mathcal{T}(X, {\phi}_{XY}) \otimes Y ) + \lambda \sum ||\nabla{\phi}_{XY}||^2,
\label{eqn:regist_loss}
\end{align}
where $\lambda$ is a hyper-parameter, ${\phi}_{XY}$ is a deformation vector field from $G_X$ with the input $X$ and $Y$, and $\otimes$ denotes the local cross-correlation, which is computed by:
{
\begin{align}
A\otimes B = \sum_{\bv \in \Omega} \frac{ \left(\sum_{\bv_i} (A(\bv_i)-\bar A(\bv))(B(\bv_i)-\bar B(\bv))\right)^2}{  \left(\sum_{\bv_i} (A(\bv_i)-\bar A(\bv))^2 \right)  \left(\sum_{\bv_i} ( B(\bv_i)-\bar B(\bv))^2\right)},
\label{eqn:cross-correlation}
\end{align}
where $\Omega$ denotes the whole 3D volume, and  $\bar A(\bv)$ and $\bar B(\bv)$ denote the local mean value of volume $A(\bv)$ and $B(\bv)$, respectively. Here, $\bv_i$ iterates over a $w\times w \times w$ pixels  around $\bv$ (or $w\times w$ for the 2-D registration case), with $w=9$ in our study.
}

\subsubsection{Cycle Loss}
\begin{figure}[t!]
\centering
\includegraphics[width=8cm]{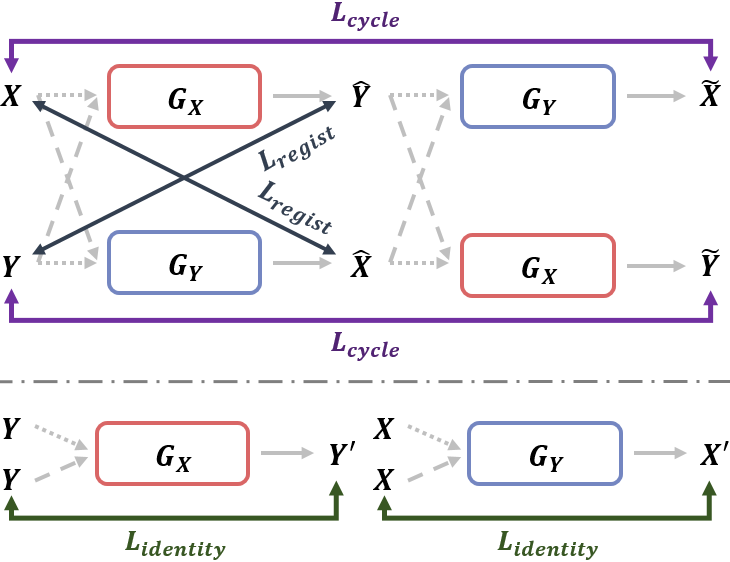}
\caption{The diagram of loss function structure in our proposed method. The registration loss function, $\mathcal{L}_{regist}$, computes dissimilarity in shape of the deformed and fixed image. The cycle loss function, $\mathcal{L}_{cycle}$, allows the displacement fields to preserve topology between the moving and deformed image. The identity loss function, $\mathcal{L}_{identity}$, enables the network to be trained stable to generate displacement vector fields.}
\label{fig:loss}
\end{figure}

To retain the topology during the deformation, we design the cycle consistency on a pixel level of images as shown in Fig.~\ref{fig:loss}. Specifically, an image $X$ is first deformed to an image $\hat Y$, after which the deformed image is registered again by another network to generate image $\tilde X$ in the proposed framework. Then, the cycle consistency is applied between the re-deformed image $\tilde X$ and its original image $X$ to impose $X\simeq \tilde X$. Similarly, an image $Y$ should be successively deformed by the two networks to generate image $\tilde Y$, and the cycle consistency allows to impose $Y\simeq \tilde Y$.

Here, one of the important parts of cycle loss for our registration framework is that the network receives two inputs: moving and fixed images. Thus, the correct implementation of the cycle consistency should be given by as the vector-form of the cycle consistency condition:
\begin{eqnarray*}
(X, Y) \simeq \left( \mathcal{T}(\hat Y,\hat {\phi}_{YX}),  \mathcal{T}(\hat X,\hat {\phi}_{XY}) \right)
\end{eqnarray*}
where
\begin{eqnarray}\label{eq:hat}
 (\hat Y, \hat X):= \left(\mathcal{T}(X, {\phi}_{XY}), \mathcal{T}(Y, {\phi}_{YX})\right) .
\end{eqnarray}
Therefore, the cycle loss is computed by:
\begin{align}
    \mathcal{L}_{cycle}&  (X,Y, G_X, G_Y) \nonumber \\
    & =\|\mathcal{T}(\hat Y,\hat {\phi}_{YX}) - X\|_1 + \|\mathcal{T}(\hat X,\hat {\phi}_{XY}) - Y\|_1, 
\label{eqn:cycle_loss}
\end{align}
where $||\cdot||_1$ denotes the $l_1$-norm.

\begin{figure*}[t!]
\centering
\includegraphics[width=\linewidth]{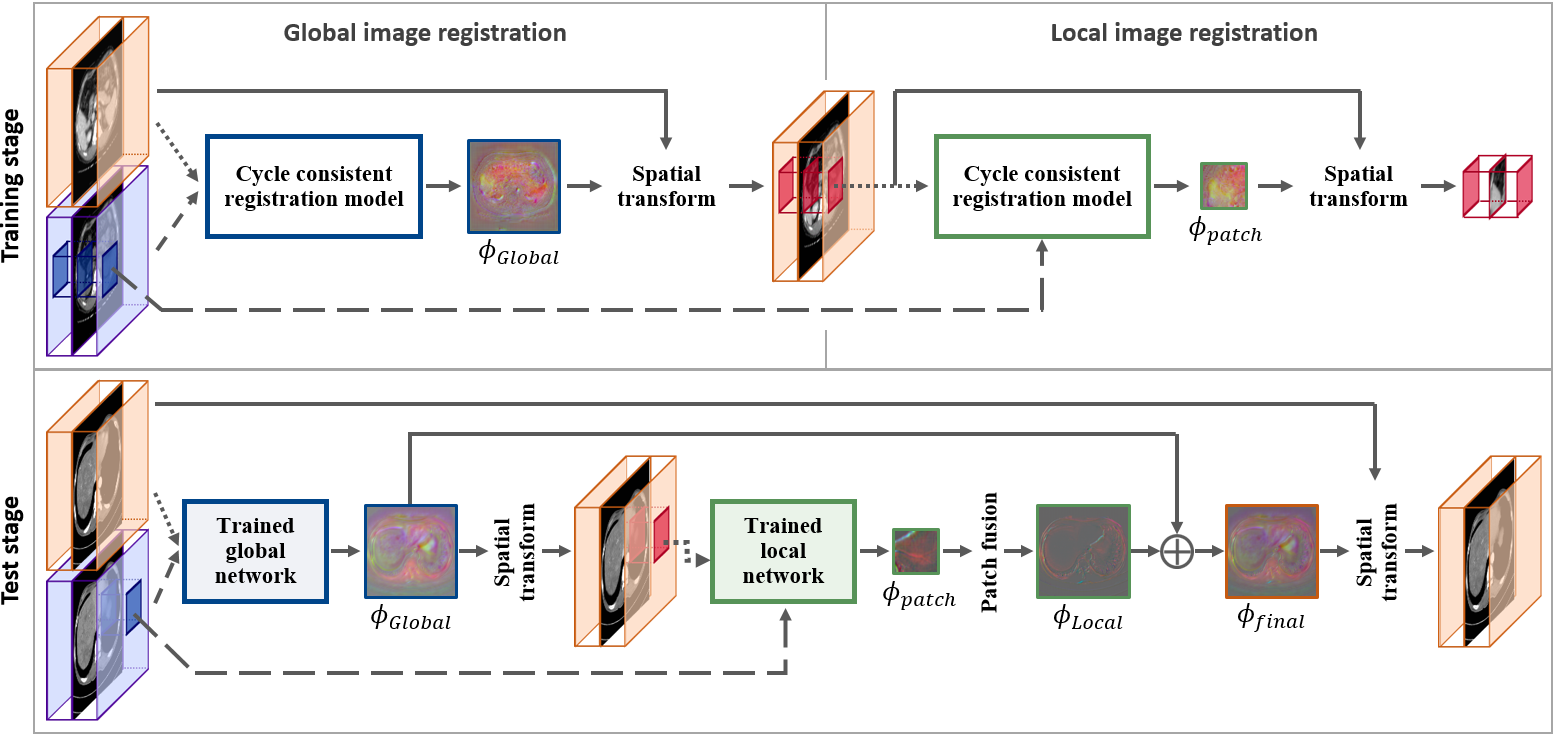}
\caption{The flow diagram of the multiscale registration method for large-scale images. The upper part illustrates the flow of training stage for global and local image registration. The lower part shows the flow of test stage using the trained global and local registration networks for a given moving and fixed images. The short- and long-dashed lines indicate moving image and fixed image, respectively.}
\label{fig:method}
\end{figure*}

\subsubsection{Identity Loss}
When deforming images by displacement vector fields, the stationary regions of images should not be changed as a fixed points. To consider this and improve the registration accuracy, as shown in Fig.~\ref{fig:loss}, we design the identity constraint by imposing that the input image should not be deformed when the identical images are used as both a moving and fixed images. We implement the identity loss as following:
\begin{align}
\mathcal{L}&_{identity} (X, Y, {G_X}, {G_Y}) \nonumber \\
=&  -(\mathcal{T}(Y, G_X(Y, Y)) \otimes Y) - (\mathcal{T}(X, G_Y(X, X)) \otimes X ),
\label{eqn:identity_loss}
\end{align}
where $\otimes$ denotes the local cross correlation defined in~\eqref{eqn:cross-correlation}. Since minimizing the negative of cross correlation loss allows the similarity between deformed image and fixed images to be maximized, the maximum for the identical inputs can be achieved by not performing deformation (or trivial identity deformation). Thus, this identity loss prevents unnecessary deformation, increasing the stability of the deformation vector field estimation in stationary regions.

\subsection{Spatial Transformation Layer}
In order to deform a moving image $X$ with the displacement vector field $\phi$ from the network, we add the spatial transform layer $\mathcal{T}$ proposed in \cite{jaderberg2015spatial} to the network. Specifically, for 3D image registration in our experiments, we adopt the 3D transformation function with trilinear interpolation, which can be defined as:
\begin{align}
& \mathcal{T}(X, \phi) \nonumber \\ 
& = \sum\nolimits_{q \in \mathcal{N}(p+\phi(p))}^{ }X(q)\prod\nolimits_{d \in \{i,j,k\}}^{ }(1-|p_d + \phi(p_d)-p_d|),
\label{eq:st}
\end{align}
where $p$ indicates the pixel index, $\mathcal{N}(p+\phi(p))$ denotes the 8-pixel cubic neighborhood around $p+\phi(p)$, and $d$ is three directions in 3D image space. Similarly, in case of 2D image registration, we deform the image by applying bilinear interpolation in the spatial transform layer. Since this grid sampling via spatial transformer network is differentiable, our deep learning model can be trained by backpropagating errors during optimization.

\subsection{Multiscale Image Registration}
Although the proposed CycleMorph provides powerful deformation on various image domains, deep neural networks should
be trained using GPU, whose bottleneck is the limited memory. Especially, this is a problem for 3D image registration,
such as contrast enhance CT registration of a liver at multiple time points.

Since CycleMorph can be applied not only to full-sized images but also to downsampled images and local patches, the issue of memory limitation can be resolved by multiscale image registration method, i.e. global registration followed by local registration. 

Fig~\ref{fig:method} shows the schematic flow diagram of training and test stages in the proposed multiscale registration method. More details are as follows.

\subsubsection{Training Stage}
In the training stage, the global and local registration networks are separately trained by the proposed cycle consistent model. Specially, the global image registration model is first trained on sub-sampled image pairs, and then  we obtain the full-resolution deformed images by up-sampling the deformation fields. Then, the local image registration model is trained on patches extracted from the deformed images from the global registration and the original fixed images.

\subsubsection{Test Stage}
Although the global and local registration networks are trained separately, the successive deformation of a moving image with two registration networks potentially reduces the registration  accuracy due to the accumulation of interpolation errors at each stage. Therefore, rather than deforming a moving image twice, the trained global and local networks are applied successively to estimate the deformation field at each scale, and the final deformation of the moving image is performed only once using the refined deformation field (see Fig.~\ref{fig:method}). 

Specifically, given a new pair of input composed of moving and fixed images, the trained global registration network generate an intermediate deformed image and the corresponding deformation vector field $\phi_{global}$. Then, the local registration network takes an input of patches extracted from the deformed image and the fixed image, so that it can generate the deformation field  $\phi_{patch}$ for each patch. By the fusion of deformation fields from all patches generated by the local network, the whole deformation field $\phi_{local}$ at the fine scale is obtained. 
Then, by adding the global and local registration fields, $\phi_{global} + \phi_{global}$, a final deformation vector field $\phi_{final}$ can be estimated. With this $\phi_{final}$ and a spatial transformer, the moving image is finally deformed once to align with the fixed target image.

Accordingly, the final deformed image can have a resolution similar to that of the original moving image without accumulating an interpolation error.

\section{Method}
\label{sec:experiment}
To demonstrate the flexibility and improved performance of the proposed method, we conducted experiments using images from various application domains. First, we apply our method to face expression images to show the registration performance on 2D images. Second, we apply our method for 3D brain MR registration benchmark data set, in which individual brain images are registered to a common atlas. Finally, we verify our approach using a very challenging registration problem with liver CECT data set, where extensive deformation  from large 3D volumes  should be estimated for multiphase contrast enhancement pattern analysis.

\subsection{Datasets}

\subsubsection{Facial Expression Image}
The 2D face expression images are obtained from Radboud Faces Database (RaFD) \cite{langner2010presentation}. This provides eight different facial expression images for each 67 subjects; neutral, angry, contemptuous, disgusted, fearful, happy, sad, and surprised. This dataset also provides three different gaze directions for all facial expressed images so that there are total 1,608 images. We divided the dataset by 53, 7, and, 7 participants for training, validation, and test images, respectively, and used all pair of face images gazing the same direction. We cropped all images to $640 \times 640$ and resized them into $128 \times 128$.

\subsubsection{Brain MRI}
For brain MR image registration task, we used OASIS-3 \cite{lamontagne2018oasis} dataset. This provides 1,249 T1-weighted 3D brain MR images and corresponding volumetric segmentation results produced through FreeSurfer \cite{fischl2012freesurfer}. Specifically, we first preprocessed the data using standard preprocessing steps: resampling all scans to $256 \times 256 \times 256$ grid with 1mm$^3$ isotropic voxels, affine spatial normalization, and brain extraction. Then, we cropped the images to $160\times 192 \times 224$, and divided by 255. We used 1027 scans for network training, 93 scans for validation, and 129 scans for test data.

\subsubsection{Multiphase Liver CT}
The multiphase liver CT scans are provided by Asan Medical Center, Seoul, South Korea. Each scans was acquired from the patients with risk factors for HCC in the liver. The scan is 4D liver CT in that 3D volumes in four-phase (unenhanced, arterial, portal and 180-s delayed phases) before and after the contrast agent injection. The data have a resolution of $512 \times 512 \times depth$, where $depth$ is the number of slices for each CT images, and the slice thickness is 5mm. We trained the networks for image registration using 555 scans and evaluated our method on 50 test scans.

\begin{figure}[h!]
\centering
\includegraphics[width=6cm]{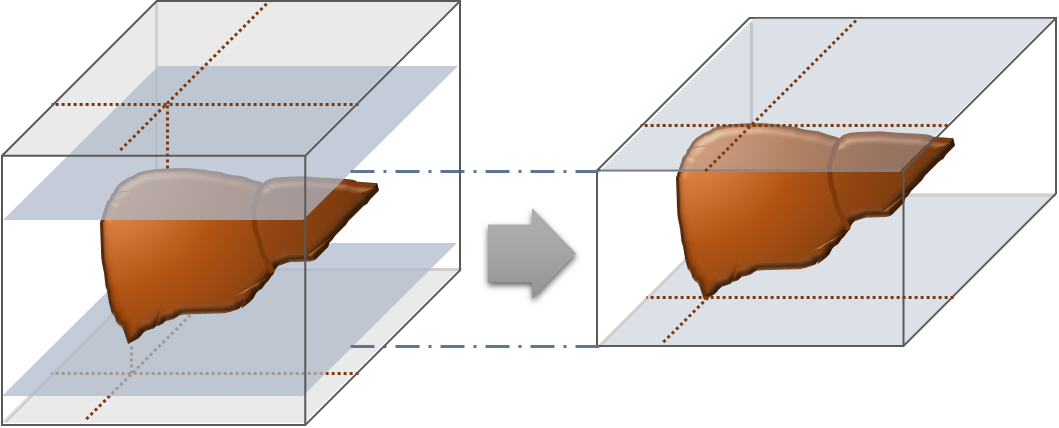}
\caption{The illustration of the process to extract slices including liver in abdominal CT images.}
\label{fig:preprocessing}
\end{figure}

Here, since $depth$ of multiphase images may be all different due to their different scanning time, image coverage, and field-of-view of images, we extracted slices including liver by a segmentation network trained by an improved U-Net  rather than resampling data into same image size. Then, to stack moving and fixed images along the channel direction as a network input, we performed zero-padding to the above and below volumes to make the number of slices same as shown in Fig.~\ref{fig:preprocessing}, which allows the input images to have same characteristics with the original images without any information loss in liver region. We normalized the images with the maximum value of each volume.

\subsection{Implementation Details}
The proposed deformable registration method was implemented in Python using pyTorch library. The specific implementation details for face and medical image registration tasks are as follows.


\subsubsection{2D Face Expression Image Registration}
For the face image registration, we employed 2D U-Net \cite{ronneberger2015u} that takes 2D images as an input of moving and fixed images and generates a deformation field in width and height directions. In training of the network, we used the Adam optimization algorithm with learning rate $4 \times 10^{-5}$ and batch size 1. We set the hyper-parameters as $\alpha=0.5$, $\beta=1$, and $\lambda=1$. We augmented the data by randomly vertical flipping and trained the model for 20 epochs using a single GPU, NVIDIA GeForce GTX 1080 Ti. For the input, we converted the RGB images to gray-scale, but to obtain deformed images with RGB channels, we applied the same deformation fields of gray-scale images to each RGB channels at the test stage.

\subsubsection{3D Medical Image Registration}
In order to evaluate the proposed model with 3D medical image registration task, we adopted 3D CNN that takes 3D volumes and generates a displacement vector field in width-, height-, and depth direction. We used VoxelMorph-1 \cite{balakrishnan2018unsupervised} as a baseline network, so that our deep learning model without both the cycle and identity loss is equivalent to VoxelMorph-1. This network architecture consists of encoder, decoder and their connections similar to U-Net \cite{ronneberger2015u}. Here, because of the high memory usage for training the 3D CNN, we set the batch size to 1. For data augmentation, we adopted random horizontal and vertical flipping and rotation with 90 degree for each training volume pair to improve registration performance without over-fitting. 
For brain MRI registration, we set the hyper-parameters as $\alpha=0.1$, $\beta=0.5$, and $\lambda=1$. To train the networks, we applied Adam with momentum optimization algorithm with learning rate $2 \times 10^{-4}$. Using a single GPU, NVIDIA Titan RTX, we trained the model for 30 epochs. 
Here, even though the brain registration task fits in the GPU memory, we also tested our multiscale registration method to compare with the existing registration approach. For training of local registration model in the multiscale approach, we extracted patches from the globally deformed image and fixed images with size of $p \times p \times p$, where $p=64$ in our experiment, and at the inference phase, we got the local registration fields by overlapping the patches by $\frac{3}{4}p \times \frac{3}{4}p \times \frac{7}{8}p$. We set the learning rate as $1 \times 10^{-4}$ and trained the model for 70 epochs.

For multiphase liver CT image registration, we adopted the multiscale registration method to address GPU memory limitation. For the global registration model, we sub-sampled the pair of input images from $512\times 512\times depth$ to $128 \times 128 \times depth$ to fit in the GPU memory size, but at the inference phase, we obtained full-resolution deformation fields by upsampling. 
Also, the training method of local registration model in multiscale approach was same with the brain registration method mentioned above. 
Using a single GPU, NVIDIA GeForce GTX 1080 Ti, we trained the global and local registration networks for 50 and 30 epochs, respectively, by Adam optimization with learning rate $10^{-4}$. We set the hyper-parameters as $\alpha=0.1$, $\beta=1$, and $\lambda=1$.

\subsection{Evaluation}
To verify the proposed method quantitatively, we evaluated the registration accuracy between the deformed and fixed images. 
First,  we use the common evaluation criterion by measuring the regularity of the deformation fields $\phi$.
This can be done by computing the percentage of non-positive values in determinant of Jacobian matrix on $\phi$, which can be defined by:
\begin{align}
    |J_\phi (\bv)| = | \nabla \phi(\bv) | \leq 0.
    \label{eqn:TRE}
\end{align}
where $\bv$ denotes the voxel location and $|\cdot|$ is the determinant of a matrix.
According to the property of Jacobian matrix, the deformation is diffeomorphic when the determinant of Jacobian matrix has all positive values,
so the percentage of the negative Jacobian indicates how much the registration is different from diffeormorphic registration.

\begin{figure*}[t!]
\centering
\includegraphics[width=\linewidth]{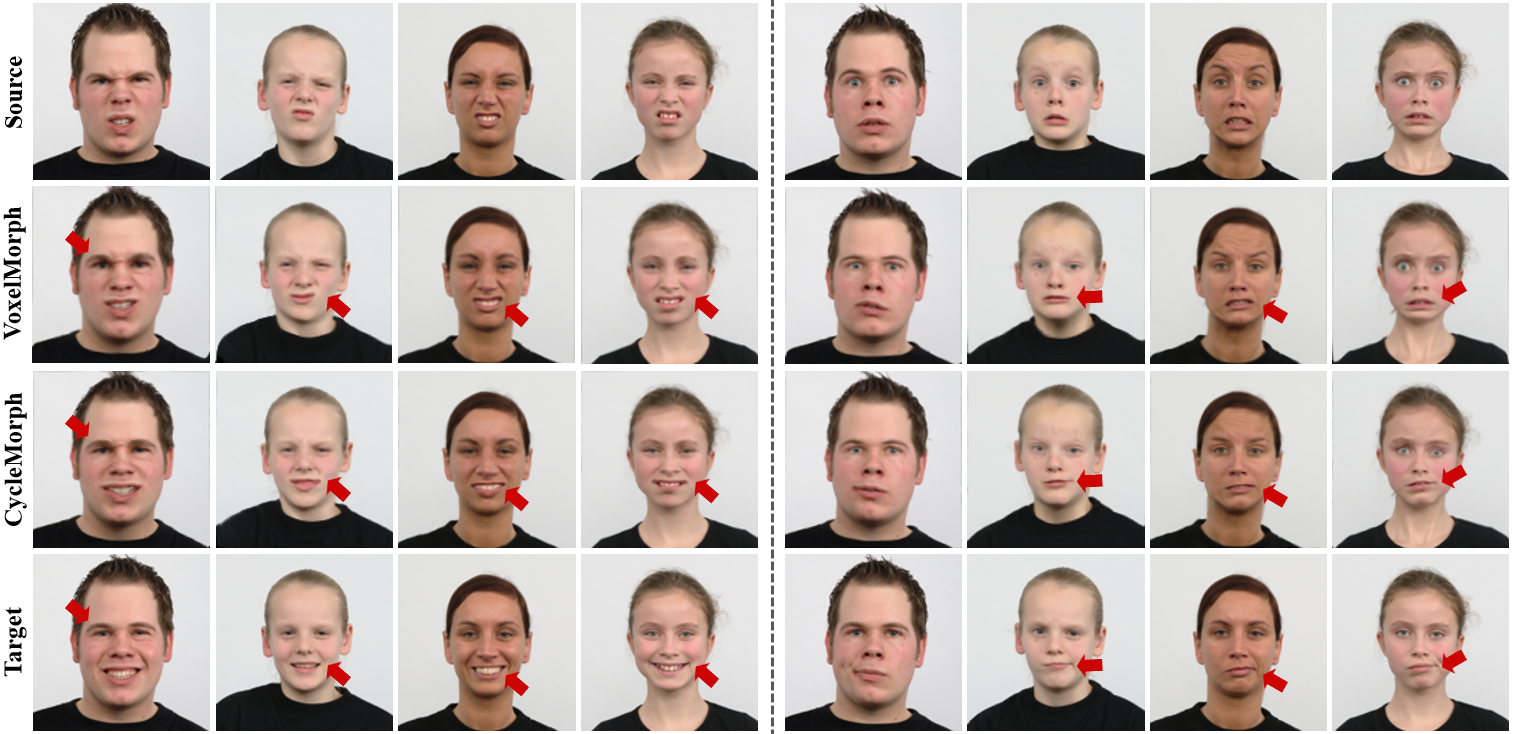}
\caption{Qualitative comparison results of face expression image registration. The red arrows indicates the remarkable parts of the results. Left: results from disgusted to happy face registration. Right: results from fearful to contemptuous face registration. First row: source (moving) image. Second row: results from VoxelMorph \cite{balakrishnan2018unsupervised}. Third row: results from our proposed CycleMorph. Fourth row: target (fixed) image.}
\label{fig:result_flag1}
\end{figure*}

Additional quantitative evaluation criterion for each datasets depends on each application: the facial expression image dataset has ground-truth labels of deformed images; the brain MR dataset has segmentation map for several brain structures; and the liver CT dataset has anatomical landmark points. Therefore, we adopted different evaluation methods for each datasets.  The details are as follows.

\subsubsection{Analysis of Face Expression Image Registration}
In the face expression image registration tasks, we deform different facial expression images of same person who gaze same direction. Accordingly,  there are ground-truth labels for all deformed images, so we evaluated the results of face image registration by the normalized mean square error (NMSE) and structural similarity (SSIM) between deformed images and fixed target images. For all pairs of face expression images, we averaged the scores for quantitative analysis.

\subsubsection{Analysis of Brain MRI Registration}
Since the brain MR dataset we used has segmentation labels for anatomical structures of brain, we evaluated the registration performance using the Dice score between the deformed segmentation map and fixed atlas segmentation label, which can be computed as:
\begin{align}
    Dice(A, B) = \frac{2 TP}{2TP + FP + FN},
    \label{eqn:dice}
\end{align}
where $TP$, $FP$, and $FN$ are the number of pixels of true positive, false positive, and false negative regions. Among the segmented anatomical structures, we extracted 30 structures that are typically composed of over 100 pixels in a volume. To get segmentation maps for the registered images, we deformed the original segmentation map of moving image with the deformation fields computed from the registration networks
between the original image and the atlas.

\subsubsection{Analysis of Liver CT Registration}
For the quantitative evaluation of liver CT registration, we computed the target registration error (TRE) on the 20 anatomical and pathological points in the liver and adjacent organs on the axial portal-phase images of the 50 test CT scans, which are marked by radiologists. The TRE can be computed by the average Euclidean distance as following:
\begin{align}
    TRE(A, B) = \frac{1}{N} \sum_{i=1}^{N} \| a_i - b_i \|,
\end{align}
where $N$ is the number of landmark points, $a_i$ and $b_i$ is the $i$-th landmark coordinate vectors in the moving image $A$ and fixed image $B$, respectively. Also, we measured differences of liver cancer size with major and minor lengths of cancer region to verify the performance in the view point of tumor diagnosis. The specific information of the marking points is described in Appendix. 

\subsubsection{Comparative Methods}
In order to verify the improved performance of  the proposed method, we employed several comparative methods that show the state-of-the-art performance in the image registration: Elastix \cite{klein2009elastix}, SyN \cite{avants2008symmetric} by Advanced Normalization Tools (ANTs) \cite{avants2011reproducible}, VoxelMorph \cite{balakrishnan2018unsupervised}, and MS-DIRNet \cite{lei20204d}. Except for the classical approach, Elastix and ANTs, we used VoxelMorph-1 proposed in \cite{balakrishnan2018unsupervised} as a baseline network, and employed same parameters for fair comparison.  Since MS-DIRNet \cite{lei20204d} is one of the representative
 multi-scale approaches to address GPU memory issues,  we used MS-DIRNet as a baseline method to compare our multiscale implementation of CycleMorph.

\section{Experimental Results}

\subsection{Face Expression Image Registration}

\subsubsection{Qualitative Evaluation}
Fig.~\ref{fig:result_flag1} shows visual comparisons of the 2D image registration results on various face expression photos of men, women, and children. We deform the source image to follow the target image. We can observe that the proposed method deforms source images to be more similar to target images compared to VoxelMorph, especially on the region of eyes and mouth. In most of  the data set,  we found that  the proposed CycleMorph provides significantly high-quality results of image registration compared to VoxelMorph.

\begin{figure}[t!]
\centering
\includegraphics[width=9cm]{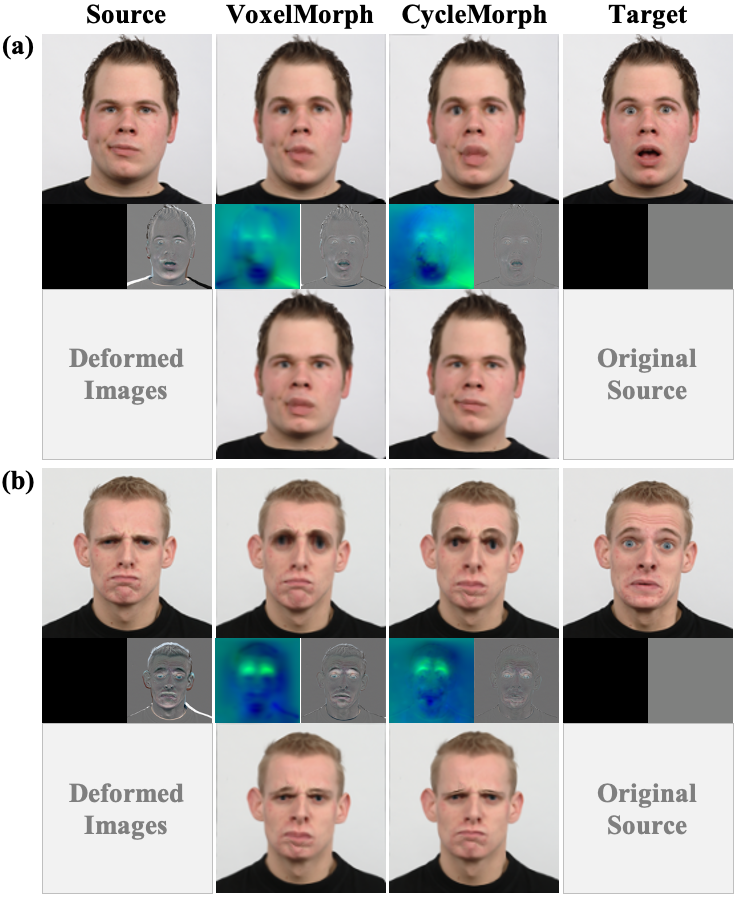}
\caption{Face expression image registration performance with various qualitative results. (a) Results from contemptuous to surprised face registration. (b) Results from angry to sad face registration. For each (a) and (b), first row shows deformed images of source into target, second row shows deformation fields (left) and difference images between the source/results and target (right) images, and third row shows re-deformed images from deformed images into the original source images.}
\label{fig:result_flag2}
\end{figure}

\begin{table}[t!]
  \caption{Quantitative evaluation results on the face expression image registration. NMSE, SSIM, and the percentage of non-positive values in determinant of Jacobian matrix of deformation fields are evaluated on the all test pairs of face expression images. (Parentheses: standard deviations across test data.)}
  \centering
 \begin{tabular}{M{1.5cm}|M{1.7cm}M{1.55cm}M{1.75cm}}
  \hline
   \multicolumn{1}{c|}{Method} & {NMSE $\scriptscriptstyle {\times 10^{-1}}$} & SSIM & $\%$ of $|J_{\phi}| \leq 0$ \\
  \hline \hline
   \multicolumn{1}{l|}{Initial}               
   & 0.356 (0.266) & 0.825 (0.065) & 0   \\ 
  \multicolumn{1}{l|}{VoxelMorph \cite{balakrishnan2018unsupervised}}
   & 0.048 (0.043) & 0.930 (0.024) & 0.053 (0.118)  \\
   \multicolumn{1}{l|}{CycleMorph}                                      
   & 0.017 (0.002) & 0.965 (0.013) & 0.016 (0.058)\\
   \hline
  \end{tabular}
  \label{table:face_quant}
\end{table}

\begin{table}[t!]
  \caption{Comparison of consistency of reverse image in the face expression image registration.
   (Parentheses: standard deviations across test data,)}
  \centering
 \begin{tabular}{M{1.5cm}|M{2cm}M{1.6cm}}
  \hline
   \multicolumn{1}{c|}{Method} & {NMSE $\scriptscriptstyle {\times 10^{-1}}$} & SSIM \\
  \hline \hline
  \multicolumn{1}{l|}{VoxelMorph \cite{balakrishnan2018unsupervised}}
   & 0.035 (0.031) & 0.932 (0.022) \\
   \multicolumn{1}{l|}{CycleMorph}                                      
   & 0.007 (0.006) & 0.976 (0.009)  \\
   \hline
  \end{tabular}
  \label{table:face_back}
\end{table}

To explicitly analyze the effect of cycle consistency to preserve diffeomorphism, we additionally performed the study on deformation fields whether the deformed images preserve topology and can be returned to their original images. In order to register the deformed images into the original images reversely, we set the forward deformed images as new source images and the original source images as new target images, and applied the same registration networks. The figures in the bottom rows of Fig.~\ref{fig:result_flag2}(a)(b) illustrate the visual comparison results of backward image registration. It shows that the proposed method provides deformed images that can be reversed to the original images, while deformed images from VoxelMorph cannot be reversed.

\begin{figure}[t!]
\centering
\includegraphics[width=\linewidth]{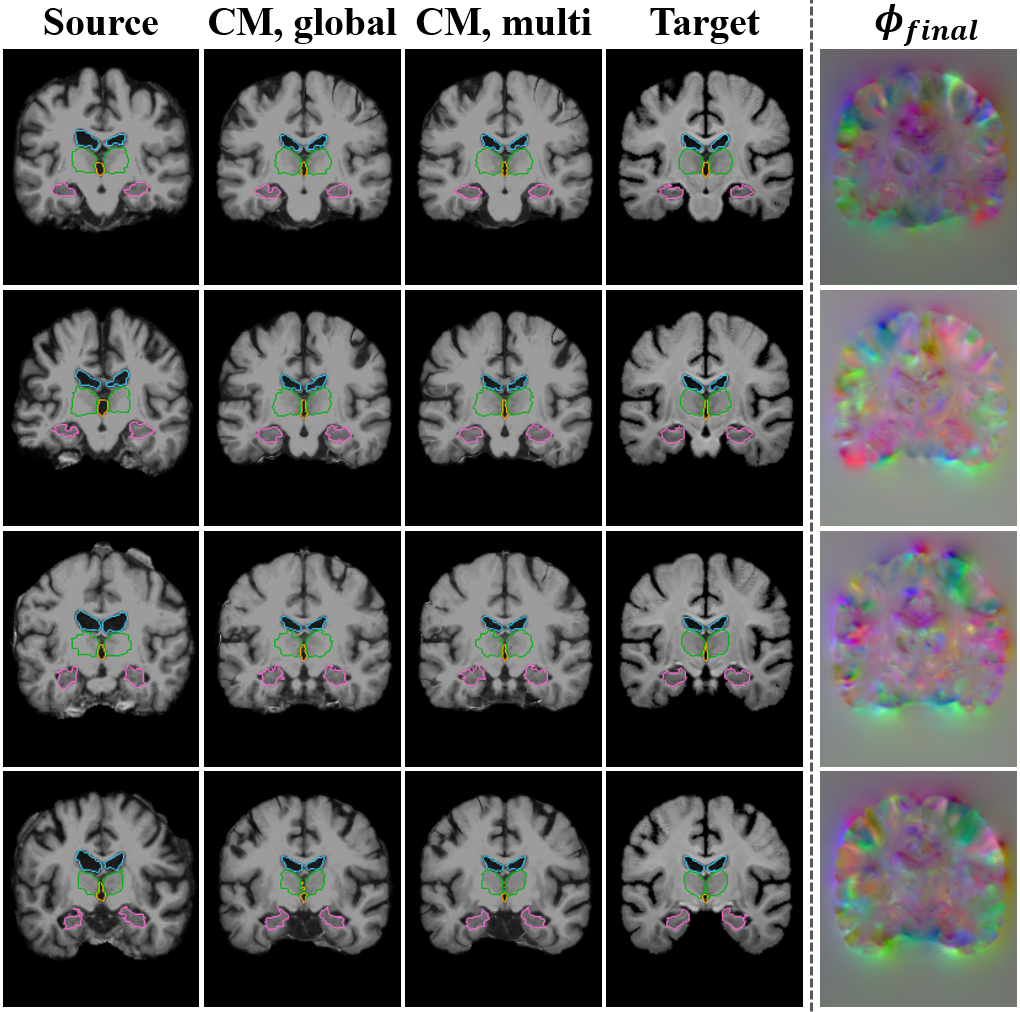}
\caption{Qualitative results of atlas-based brain MR image registration of the proposed method. We overlaid boundaries of several anatomical structures (blue: ventricles, orange: third ventricle, green: thalami, pink: hippocampi). The moving source images are in first column, deformed images from the proposed CycleMorph (CM) are in second column (global) and third column (multiscale), and the fixed target images are in fourth column. The last column shows the corresponding deformation fields $\phi_{final}$ to the warped images of our multiscale registration method.}
\label{fig:result_brain1}
\end{figure}

\begin{figure*}[t!]
\centering
\includegraphics[width=\linewidth]{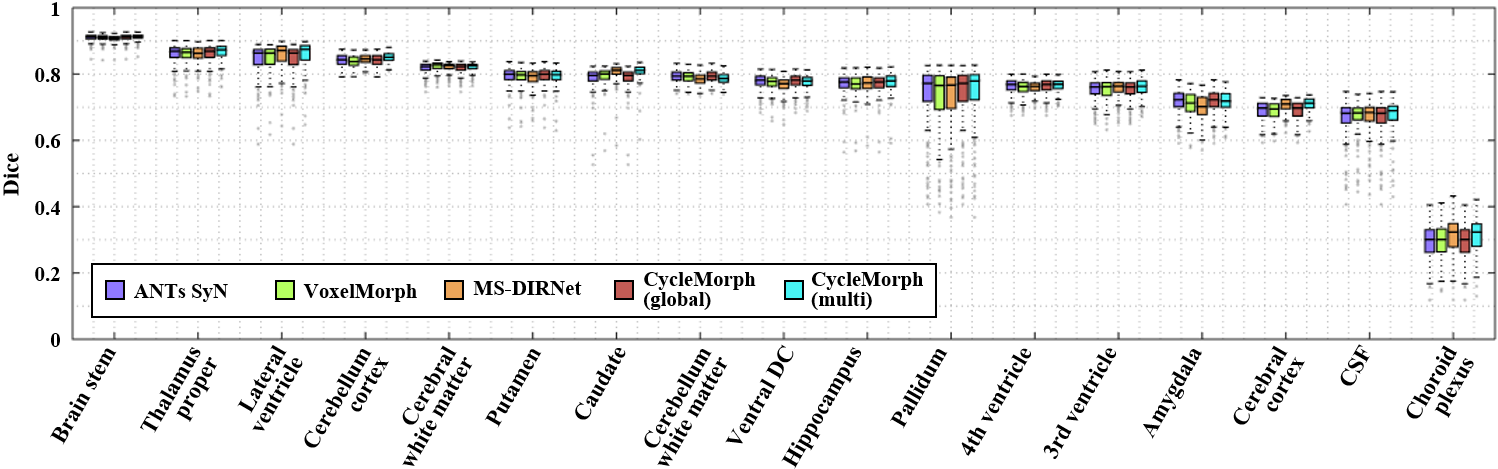}
\caption{Dice scores on the deformed segmentation maps of brain anatomical structures for quantitative comparisons of atlas-based brain MR image registration.}
\label{fig:result_brain2}
\end{figure*}

\subsubsection{Quantitative Evaluation}
Table~\ref{table:face_quant} includes the quantitative evaluation results of the comparative method (VoxelMorph) and our proposed method, which shows that our CycleMorph significantly outperforms VoxelMorph in all metrics. To compare the performance of image registration effectively, we computed evaluation scores on the source and target images before the registration. By comparison, we found that the proposed CycleMorph decreases NMSE by 0.034 and increases SSIM by 0.140 compared to the initial. Also, our method outperforms VoxelMorph by 0.037 \% gain in the metric on Jacobian determinant. 

Additionally, Table~\ref{table:face_back} shows the NMSE and SSIM between the re-deformed images (the figures in the bottom rows of Fig.~\ref{fig:result_flag2}(a)(b) and their original moving images. The reversed images from our method is very similar to the original images with lower NMSE and higher SSIM compared to VoxelMorph. Therefore, we can confirm that the cycle constraint in our proposed method plays an important role in producing deformation fields that guarantee topology preservation of input images  with less folding problem.

\subsection{Brain MR Image Registration}

\subsubsection{Qualitative Evaluation}
The results of atlas-based brain MR image registration are shown in Fig.~\ref{fig:result_brain1}. The proposed CycleMorph method deforms images accurately for each pairs of the moving source and fixed target images, which can be specifically verified with the segmentation boundaries of several brain structures. 
In addition, thanks to the cycle constraint, we can confirm that the image registration is performed by the smooth deformation fields without singularities.

\subsubsection{Quantitative Evaluation}
To evaluate the proposed method on atlas-based brain MR image registration, we compared the method to several comparative methods: ANTs for traditional method, VoxelMorph and MS-DIRNet for deep-learning-based global and multiscale approaches. Fig.~\ref{fig:result_brain2} represents Dice scores for the evaluated anatomical structures across test scans. The scores of left and right brain structures are averaged into one score. For all structures, our CycleMorph models achieve higher scores than VoxelMorph in global registration and MS-DIRNet in multiscale registration. In particular, on some structures such as brain stem, thalamus proper, hippocampi, pallidum, and forth ventricle, our global and multiscale CycleMorph models perform better than the comparative methods.

Table~\ref{table:brain_quant} shows the quantitative evaluation results with average Dice scores across all structures and scans, the percentage of non-positive values in Jacobian determinant, and runtime. As for the global registration, the proposed CycleMorph shows higher Dice measures with less percentage of non-positive Jacobian determinant compared to VoxelMorph.  Thus, we can confirm that the CycleMorph enforces the diffeomorphic deformations and performs effective and accurate 3D image registration.  
These results are similarly shown in the comparison of multiscale registration methods with MS-DIRNet and our method. 
However, in both MS-DIRNet and our method,
the Jacobian determinant index became inferior with multiscale registration.

\begin{table}[t!]
  \caption{Quantitative evaluation results on the brain MR image registration. Dice, the percentage of non-positive values in determinant of Jacobian matrix of deformation fields, and runtime (min) are computed on the all test scans. (Parentheses: standard deviations across test data.)}
  \centering
 \begin{tabular}{M{1.3cm}|M{1.6cm}M{1.75cm}M{1.4cm}}
  \hline
   \multicolumn{1}{c|}{Method} & Dice & $\%$ of $|J_{\phi}| \leq 0$ & Time \\
  \hline \hline
   \multicolumn{1}{l|}{Initial}               
   & 0.616 (0.171) & 0  & 0   \\ 
   \multicolumn{1}{l|}{ANTs SyN \cite{avants2008symmetric}}               
   & 0.752 (0.140) & 0.400 (0.100) & 122 (CPU)   \\ 
  \multicolumn{1}{l|}{VoxelMorph \cite{balakrishnan2018unsupervised}}
   & 0.749 (0.145) & 0.553 (0.075) & 0.01 (GPU)  \\
   \multicolumn{1}{l|}{MS-DIRNet \cite{lei20204d}}
   & 0.751 (0.142) & 0.804 (0.089) & 2.06 (GPU)  \\
   \multicolumn{1}{l|}{CycleMorph, global}                                      
   & 0.750 (0.144) & 0.510 (0.087) & 0.01 (GPU)\\
   \multicolumn{1}{l|}{CycleMorph, multi}                                      
   & 0.756 (0.141) & 0.788 (0.100) & 2.18 (GPU)\\
   \hline
  \end{tabular}
  \label{table:brain_quant}
\end{table}

\begin{table}[t!]
  \caption{Results of study on local patch size. Dice and the percentage of non-positive values in determinant of Jacobian matrix of deformation fields are computed on the all test scans. "p\#" denotes patch size $\# \times \# \times \#$ used in local image registration. (Parentheses: standard deviations across test data.)}
  \centering
 \begin{tabular}{M{2.2cm}|M{2.0cm}M{2.0cm}}
  \hline
   \multicolumn{1}{c|}{Method (CycleMorph)} & Dice & $\%$ of $|J_{\phi}| \leq 0$ \\
  \hline \hline
   \multicolumn{1}{l|}{global}                                      
   & 0.7502 (0.144) & 0.510 (0.087) \\
   \multicolumn{1}{l|}{global + local(p64)}                                      
   & 0.7564 (0.141) & 0.788 (0.100) \\
    \multicolumn{1}{l|}{global + local(p80)}                                      
   & 0.7553 (0.141) & 0.789 (0.099) \\
    \multicolumn{1}{l|}{global + local(p96)}                                      
   & 0.7543 (0.141) & 0.855 (0.098) \\
   \hline
  \end{tabular}
  \label{table:brain_patch}
\end{table}

\subsubsection{Study on Local Patch Size}
Since the patch size in local image registration of our method can be various, we also studied on the effect of local patch size in the proposed model. As shown in Table~\ref{table:brain_patch}, we set the global registration results as a baseline, and conducted the experiment of local registration with different patch sizes. 

When we compared the results with Dice scores and Jacobian determinant, all multiscale methods improve the global registration results on Dice score but generate deformation fields with more non-positive values in Jacobian determinant. Also, when the patch size is smaller, we can observe that the result on Dice score of anatomical structures tends to be higher, and the deformation regularity is also better with less folding problem. From these results, we extract patch with $64\times 64\times 64$ in our experiments.

\begin{figure*}[t!]
\centering
\includegraphics[width=\linewidth]{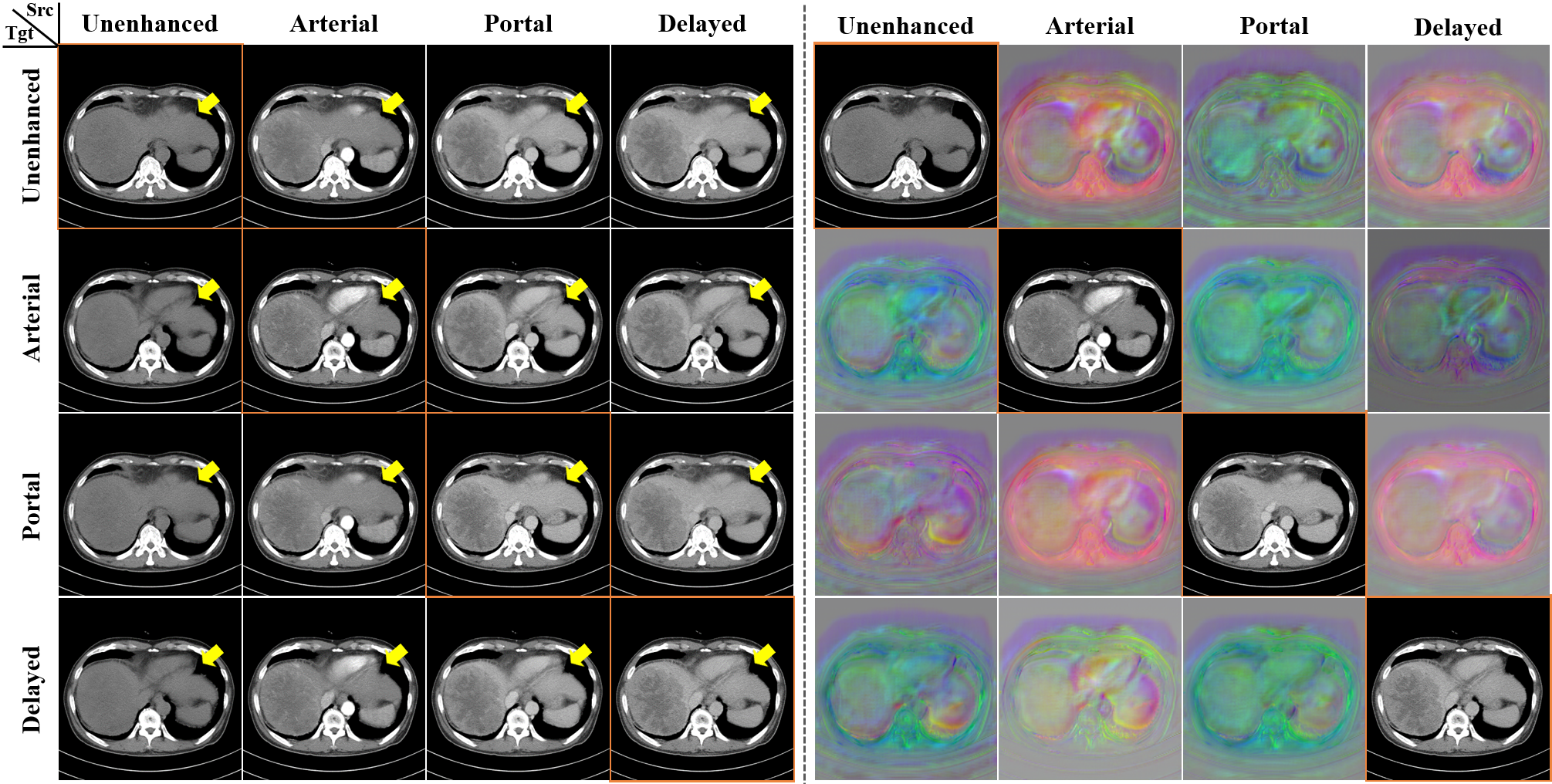}
\caption{Qualitative results of multiphase liver CT registration with a single trained network. Left: deformed images from the source (src) to target (tgt) images. Right: the deformation vector fields for the left deformed results. The diagonal images with orange box are original images, and they are deformed to other phase images as indicated by each row. The $(i, j), i \neq j$, element of the figure represents the deformed image to the $i$-th phase from the $j$-th phase original image. The yellow arrows with the same position indicate the remarkable parts of the results. }
\label{fig:result_asan1}
\end{figure*}

\begin{figure}[t!]
\centering
\includegraphics[width=\linewidth]{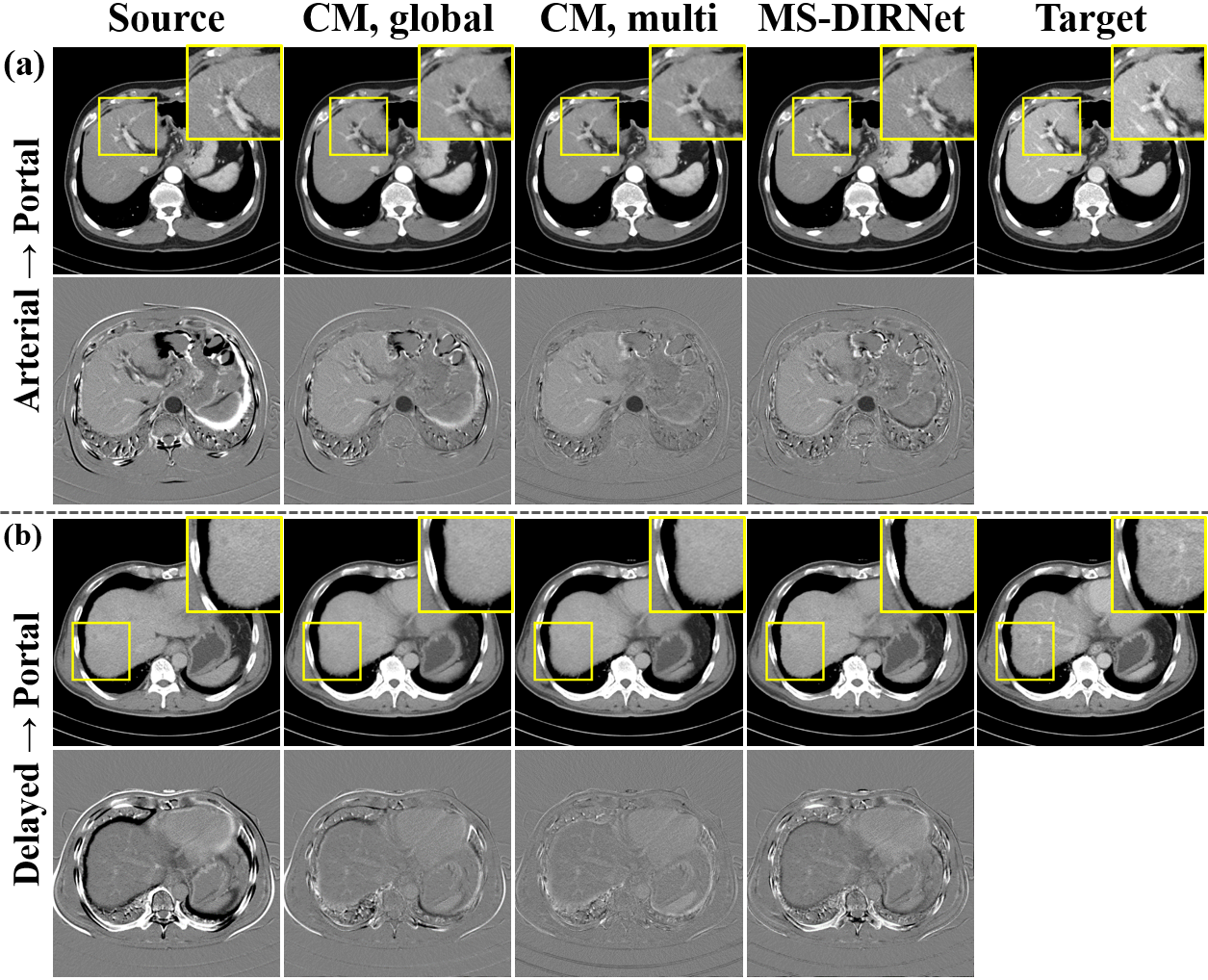}
\caption{Global and multiscale (global followed by local) registration results of the proposed CycleMorph (CM) on the multiphase liver CT dataset. (a) Results from the images in arterial to portal phases. (b) Results from the images in delayed to portal phases. For each (a) and (b), the yellow box shows the remarkable parts, and second row shows difference images between the source/results and target images.}
\label{fig:result_asan2}
\end{figure}

\subsection{Multiphase Liver CT Image Registration}

\subsubsection{Qualitative Evaluation}
Fig.~\ref{fig:result_asan1} and Fig.~\ref{fig:result_asan2} illustrate the multiscale registration results by CycleMorph. Specifically, Fig.~\ref{fig:result_asan1} shows that the proposed method provides accurate registration results with smooth deformation vector fields on the all multiphase 3D images with different contrast. In addition, Fig.~\ref{fig:result_asan2} shows that the multiscale registration performance is improved over the global registration results, which is well visualized in the difference images between the deformed images and target images. From this result, we can confirm that the global registration tends to deform whole shape of the source images to fit into the fixed target images, while the local registration provides the local region deformation.

\begin{table*}[t!]
  \caption{Quantitative evaluation results on the multiphase liver CT image registration. TRE ($mm$), tumor size differences, and the average test time (min) are evaluated on the deformed arterial/delayed images into the fixed portal image. (Parentheses: standard deviations across subjects.)}
  \centering
 \begin{tabular}{M{2.4cm}|M{1.7cm}|M{1cm} M{1cm}| M{1.5cm}|M{1.7cm}|M{1cm} M{1cm}| M{1.5cm}}
  \hline
  \multirow{3}{*} {{Method}} & \multicolumn{4}{c|}{\textbf{Arterial $\rightarrow$ Portal}} & \multicolumn{4}{c}{\textbf{Delayed $\rightarrow$ Portal}}\\ 
  \cline{2-9}
  & \multirow{2}{*}{TRE} & \multicolumn{2}{c|}{Tumor size diff} & \multirow{2}{*}{Time}  & \multirow{2}{*}{TRE} & \multicolumn{2}{c|}{Tumor size diff} & \multirow{2}{*}{Time} \\
  & & {Major} & {Minor} &  & & {Major} & {Minor} & \\
  \hline \hline
   \multicolumn{1}{l|}{Elastix \cite{klein2009elastix}}               
   & 3.261 (1.143) & 0.981 & 0.610 & 19.64 (CPU) & 2.963 (0.913) & 0.910 & 0.577 & 19.64 (CPU) \\ 
  \multicolumn{1}{l|}{VoxelMorph \cite{balakrishnan2018unsupervised}} 
   & 6.674 (4.217) & 0.789 & 1.638  & 0.18 (GPU)  & 5.351 (1.892) & 0.610 & 0.868 & 0.20 (GPU)  \\
  \multicolumn{1}{l|}{MS-DIRNet \cite{lei20204d}}                     
   & 5.021 (4.175) & 0.186 & 0.178 & 0.69 (GPU)   & 4.042 (1.938) & 0.136 & 0.102 & 0.69 (GPU) \\
    \multicolumn{1}{l|}{CycleMorph, global}                                      
   & 4.722 (3.294) & 0.631 & 0.563  & 0.06 (GPU)  & 3.902 (1.694) & 0.275 & 0.209 & 0.06 (GPU) \\
   \multicolumn{1}{l|}{CycleMorph, multi}                                      
   & 4.720 (3.275) & 0.678 & 0.607  & 0.69 (GPU)  & 3.928 (1.696) & 0.293 & 0.219 & 0.69 (GPU) \\
   \hline
  \end{tabular}
  \label{table:asan_quant}
\end{table*}

\begin{table*}[t!]
  \caption{Results of ablation study on loss function. TRE ($mm$) and the percentage of non-positive values in determinant of Jacobian matrix of deformation fields are computed on the deformed arterial/delayed images into the fixed portal image.  (Parentheses: standard deviations across subjects.)}
  \centering
 \begin{tabular}{M{4.3cm}|M{2cm}M{2.2cm}|M{2cm} M{2.2cm}}
  \hline
  \multirow{2}{*} {Method} & \multicolumn{2}{c|}{\textbf{Arterial $\rightarrow$ Portal}} & \multicolumn{2}{c}{\textbf{Delayed $\rightarrow$ Portal}} \\ 
  \cline{2-5}
  & {TRE} & $\%$ of $|J_{\phi}| \leq 0$  & {TRE} & $\%$ of $|J_{\phi}| \leq 0$ \\
  \hline \hline
   \multicolumn{1}{l|}{Proposed w/o $\mathcal{L}_{cycle}$+$\mathcal{L}_{identity}$}               
   & 5.377 (3.888) & 0.058 (0.170) & 4.415 (1.831) & 0.039 (0.064) \\ 
  \multicolumn{1}{l|}{Proposed w/o $\mathcal{L}_{cycle}$} 
   & 5.241 (4.017) & 0.085 (0.217) & 4.210 (1.737) & 0.083 (0.176) \\
  \multicolumn{1}{l|}{Proposed w/o $\mathcal{L}_{identity}$}                     
   & 5.006 (3.864) & 0.049 (0.131) & 4.212 (1.931) & 0.049 (0.117) \\
   \multicolumn{1}{l|}{Proposed (CycleMorph)}                                      
   & 4.722 (3.294) & 0.032 (0.099) & 3.902 (1.694) & 0.029 (0.084) \\
   \hline
  \end{tabular}
  \label{table:asan_ablation}
\end{table*}

\subsubsection{Quantitative Evaluation}
We performed quantitative evaluation of the registration results on the deformed images in arterial/delayed phases into portal phase that is often used as a standard in the clinical practice. Here, since the images in unenhanced phase are difficult to obtain the landmark points, we did not compute the evaluation metrics on the unenhanced phase images. Table~\ref{table:asan_quant} shows the results of average TRE, tumor size differences, and runtime for a 3D image registration with various comparative methods. 

Specifically, we can observe that the proposed method achieves significant improvement of registration performance compared to the existing deep learning methods of VoxelMorph and MS-DIRNet, while the TRE of the proposed method is slightly higher than Elastix. In particular, the tumor size differences between the deformed image and the portal phase image from our method are smaller than the comparative methods, which confirms that the proposed method provides the most accurate deformation even on the small cancer region. Although MS-DIRNet shows the smallest tumor size differences among comparisons, we can confirm that the registration quality of our CycleMorph are much better than MS-DIRNet as shown in Fig.~\ref{fig:result_asan2}. Furthermore, when we calculated the runtime of image registration, the deep learning based models takes less than 1 minutes with a single GPU, whereas the conventional method of Elastix takes approximately 20 minutes. Here, the  global registration of the proposed method only takes about 4 seconds, and the total runtime of multiscale registration is 41 seconds.

\subsubsection{Ablation Study on Loss Function}
\label{subsubsec:ablation}
To verify the effect of cycle constraint in our designed loss function, we also performed an ablation study on liver CT data by excluding the cycle loss and/or identity loss. For this study, we analyzed results of the global image registration with the same training and test procedure for fair comparison. Table~\ref{table:asan_ablation} shows the percentage of the number of non-positive values in a determinant of Jacobian matrix on deformation fields as well as TRE, which demonstrates that the registration performance change according to the loss function is remarkable. The network only trained by the registration loss, i.e. without $\mathcal{L}_{cycle}$ and $\mathcal{L}_{identity}$, deforms images with the largest errors among the methods. And both of the cycle and identity loss functions increase the accuracy of the registration.

On the other hand, the evaluation metric of Jacobian matrix emphasizes the effect of cycle consistency. Specifically, the proposed method without the cycle loss produces deformation fields with more non-positive voxels of determinant of Jacobian matrix than the proposed method with the cycle constraint. Here, the reason that the method only with the registration loss has smaller percentage of non-positive values of Jacobian determinants than the other ablated methods is because the network provides registration fields that hardly deform on the large-scale images, which can be confirmed with TRE values. In contrast, thanks to the cycle loss, the proposed method is less prone to folding problem and enhances topological preservation on 3D image registration.

\section{Conclusion}
\label{sec:conclusion}
In this paper, we presented a CycleMorph, a novel cycle consistent deep learning model for unsupervised deformable image registration method. CycleMorph imposes cycle consistency between a pair of images. Once the networks are trained, a single network can provide accurate image registration with any pair of new data. CycleMorph was also extended to the multiscale approach to deal with large volume registration problem. Experiments using the various image datasets confirmed that CycleMorph provides topology-preserved image deformation for any image pairs and provides significant performance improvement.



\appendix[Quantitative Evaluation of Multiphase Liver CT Image Registration]
\label{appen:Quantitative}
For the quantitative evaluation of registration performance on multiphase liver CT data, the marking of anatomical and pathological points and tumor size measurement were performed by an expert. Specifically, an abdominal radiologist (D.H.K., with 8-year experience with liver CT) marked the following anatomical and pathological points in the liver and adjacent organs on the axial portal-phase images of the 50 CT datasets using a dedicated software (Medical Imaging Interaction Toolkit Workbench): (1) the uppermost point of the liver (i.e., right hepatic dome); (2) the left end of the left lateral hepatic section; (3) the inferior tip of the right hepatic lobe; (4) the innermost border of the caudate lobe of the liver; (5) the most caudal part of the gallbladder fundus; (6) gallbladder stones if present; (7) the points where right, middle, left, and right inferior (if present) hepatic veins meet the inferior vena cava; (8) suprahepatic inferior vena cava at the level that shows its maximum width; (9) the caudal boundary of the splenic vein’s entry into the main portal vein; (10) the caudal boundary of the branching-off of the left portal vein from the main portal; (11) the origins of P2, P3, and P4 portal branches from the left portal vein; (12) the point of right portal vein’s branching into anterior and posterior segmental portal veins; (13) the points of right anterior and right posterior portal veins’ branching into segmental portal veins; (14) fissure for ligamentum teres or recanalized umbilical vein; (15) fissure for ligamentum venosum (at the level that shows the umbilical segment of the left portal vein); and (16) hepatic cysts or calcifications if present. In addition, the long- and short-diameters of hepatic HCCs (in the larges lesion if multiple nodules were present) were measured.



\section*{Acknowledgment}
The authors would like to thank Junyoung Kim for the discussion on the face expression dataset.

\ifCLASSOPTIONcaptionsoff
  \newpage
\fi



%
\bibliographystyle{IEEEtran}


%







\end{document}